\documentclass[conference]{IEEEtran}
\IEEEoverridecommandlockouts

\usepackage{cite}
\usepackage{amsmath,amssymb,amsfonts}
\usepackage{algorithm}
\usepackage{algpseudocode}
\usepackage{graphicx}
\usepackage{textcomp}
\usepackage{xcolor}
\def\BibTeX{{\rm B\kern-.05em{\sc i\kern-.025em b}\kern-.08em
    T\kern-.1667em\lower.7ex\hbox{E}\kern-.125emX}}

\begin{document}

\title{Incremental Gaussian Mixture Clustering for Data Streams}

\author{\IEEEauthorblockN{Aniket Bhanderi}
 \IEEEauthorblockA{\textit{Department of Computer Science} \\
\textit{University of Cincinnati}\\
Cincinnati, USA \\
bhandead@mail.uc.edu}
\and
\IEEEauthorblockN{Raj Bhatnagar}
 \IEEEauthorblockA{\textit{Department of Computer Science} \\
\textit{University of Cincinnati}\\
Cincinnati, USA \\
bhatnark@ucmail.uc.edu}
}

\maketitle

\begin{abstract}
The problem of analyzing data streams of very large volumes is important and is very desirable for many application domains. In this paper we present and demonstrate effective working of an algorithm to find clusters and anomalous data points in a streaming datasets. Entropy minimization is used as a criterion for defining and updating clusters formed from a streaming dataset. As the clusters are formed we also identify anomalous datapoints that show up far away from all known clusters. With a number of 2-D datasets we demonstrate the effectiveness of discovering the clusters and also identifying anomalous data points.
\end{abstract}
\makeatletter
\def\ps@IEEEtitlepagestyle{%
  \def\@oddfoot{\mycopyrightnotice}%
  \def\@evenfoot{}%
}
\def\mycopyrightnotice{%
\begin{minipage}{\textwidth}
  {\footnotesize © 2024 IEEE.  Personal use of this material is permitted. Permission from IEEE must be obtained for all other uses, in any current or future media, \\ including reprinting/republishing this material for advertising or promotional purposes, creating new collective works, for resale or redistribution to servers \\
  or lists, or reuse of any copyrighted component of this work in other works.\hfill}
    \end{minipage}
  \gdef\mycopyrightnotice{}
}

\section{Introduction}
\label{SectIntroduction}

Analysis and mining of streaming data requires that all analysis be performed in a single pass through the data stream. Continuous monitoring of industrial, business, and environmental systems are some applications that demand such capability. Streaming data for many domains can be typically modeled as a dynamic type of data where chunks of data arrive at successive time points. The analysis algorithms must process each data chunk, appropriately update the summaries for the data stream's evolving patterns, store these summaries in the limited sketch memory reserved for the data stream, and then discard the actual data. 

Clustering and anomaly detection are two of the important tasks of mining data streams, and often we need to perform them simultaneously. A number of algorithms have been developed to identify different types of clusters \cite{zubarouglu2021data,silva2013data}, and also anomalies \cite{chandola2009anomaly,wang2020anomaly} in data streams. These algorithms predominantly belong to partitional clustering paradigm, \cite{ailon2009streaming,braverman2011streaming,chauhan2015review,de2011extending,guha2016clustering,ordonez2003clustering,wang2020anomaly} mainly driven by the spirit of k-means algorithm, or belong to density based clustering paradigm, \cite{chandola2009anomaly,amini2014density} driven by the spirit of the DBSCAN algorithm. Some work also exists for identifying Gaussian mixtures based clusters in data streams \cite{raghunathan2017learning,song2005highly,wan2018icgt}. Analysis of data streams has many challenges that are not encountered for processing data when all of it is available simultaneously. These challenges include: 
\begin{enumerate}
\item Continuous evolution of clusters' characteristics as more data arrives. New clusters may appear, shapes and densities of older clusters may change, and some clusters may need to be merged or split.
\item Concept drift whereby some clusters exist in a data stream for some part of the time, then go missing for some time, and then reappear at later times. 
\item Anomalous nature of a data point also evolves with time. Some data points that may appear anomalous at first may not remain so after more data in their neighborhoods arrives. And some data points that do not appear to be anomalous at first, may become anomalous when new arriving data is placed much farther from them.
\end{enumerate} 

Many of the clustering algorithms use tree structures of micro-clusters \cite{aggarwal2003framework,kranen2011clustree,wan2018icgt} and perform aggregations of cluster features when at any time a snapshot of current clustering in the stream is needed. Some of these algorithms also mark each microcluster with a timestamp so that older data can be phased out from the results \cite{kranen2011clustree}. There are many real-world applications where spherical clusters generated by k-means type of algorithms are not suitable and we must capture more precise details of clusters' nature, preferably in the form of a mixture of multivariate Gaussian distributions. 
 
In this paper we present our methodology for identifying Gaussian clusters in data streams assuming that if the stream data were to be collected and analyzed together, it would be well represented by a mixture of Gaussian distributions. As the chunks of the data stream arrive we maintain and update profiles of existing clusters and also initiate new clusters as needed. Also, some data points in a newly arriving data chunks may be anomalous and we test this by computing their Mahalanobis distances from all the existing cluster profiles. We maintain summaries for this data stream in which we retain:
(i) Number of data points in each cluster 
(ii) Covariance matrix of each cluster 
(iii) Centroid of each cluster, and (iv) Coordinates of all suspected anomalous data points encountered so far in the stream. Our approach differs from some others which maintain in their Sketch memory a BIRCH like \cite{aggarwal2003framework,kranen2011clustree,wan2018icgt} tree structure of cluster features. We maintain cluster profiles for a number of clusters that is much larger than the number of clusters a user may want to see in the data stream. 
This lets us compress the cluster profiles into the desired smaller number of clusters. Our approach enables us to maintain full covariance matrix for each cluster signatures in the sketch. 

The work in \cite{wan2018icgt} seeks to perform Gaussian mixtures type of clustering, uses a tree structure of micro-clusters to store cluster features, but is limited to processing only purely diagonal covariance matrices. Managing combination of clusters features within a tree becomes challenging when full covariance matrices are included as parts of cluster features. Our approach proposed here is able to represent full Gaussian distribution details of each cluster and is therefore able to capture data's clusters more precisely. Decisions need to be made about combining newly arrived Gaussian clusters or existing smaller component clusters with other existing clusters. In traditional GMM algorithms clustering decisions for entire data are made in such a way that the Bayesian Information Criterion (BIC) is minimized, and complexity of the clustering is kept minimum. We adopt an alternate computational strategy to accomplish the same clustering objective. Minimizing entropy of the clusters also results in clustering with minimum model complexity, and it is easier to compute than the BIC. We seek to keep the entropy values of the evolving clusters to a minimum while making clustering or merging decisions. For validation we show that our resulting clusters are very close in structure to the ones that are obtained by a GMM algorithm working on the whole data simultaneously.

We show our methodology for maintaining the evolving profiles of clusters and anomalous points and demonstrate results with two publicly available datasets \cite{ClusteringDatasets}.  These results show that results of our methodology closely resemble those that would have been obtained if all streaming data were analyzed at once.

\begin{figure}[h]
    \centering
    \includegraphics[width=0.4\textwidth]{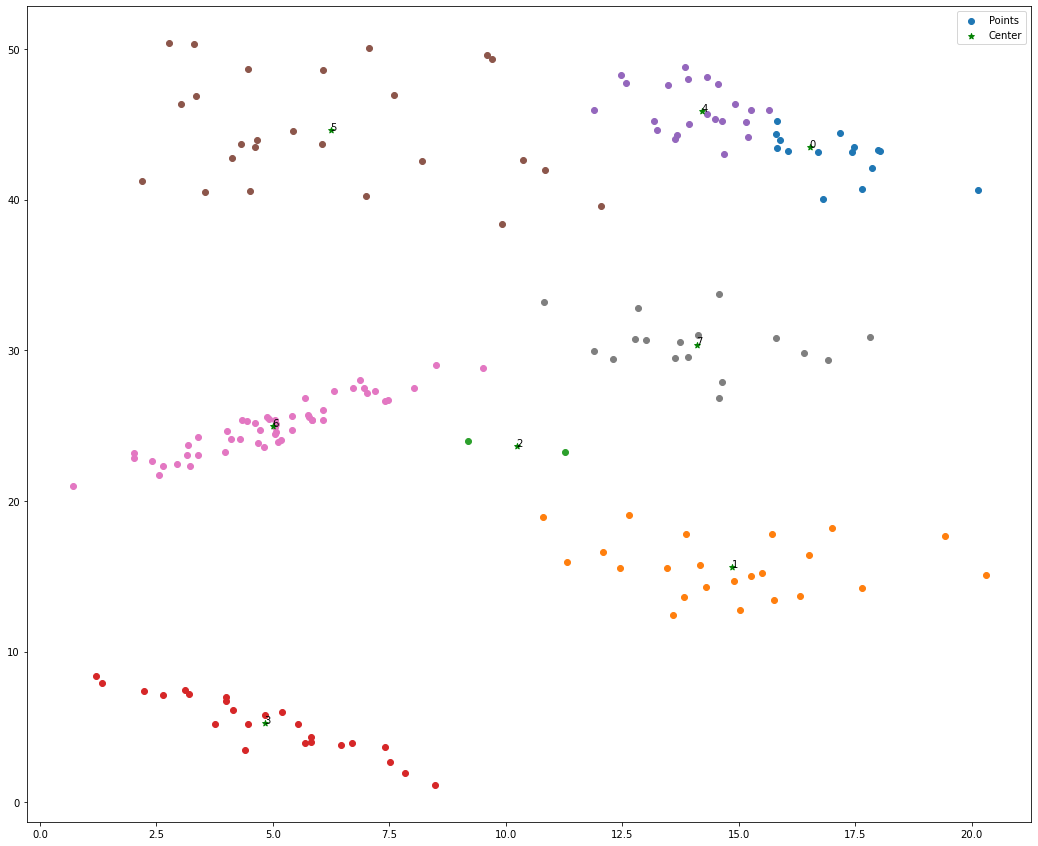}
    \caption{Incoming k$^{th}$ data chunk in Synthetic data}
    \label{fig:datachunk1}
\end{figure}

\begin{figure}[h]
    \centering
    \includegraphics[width=0.4\textwidth]{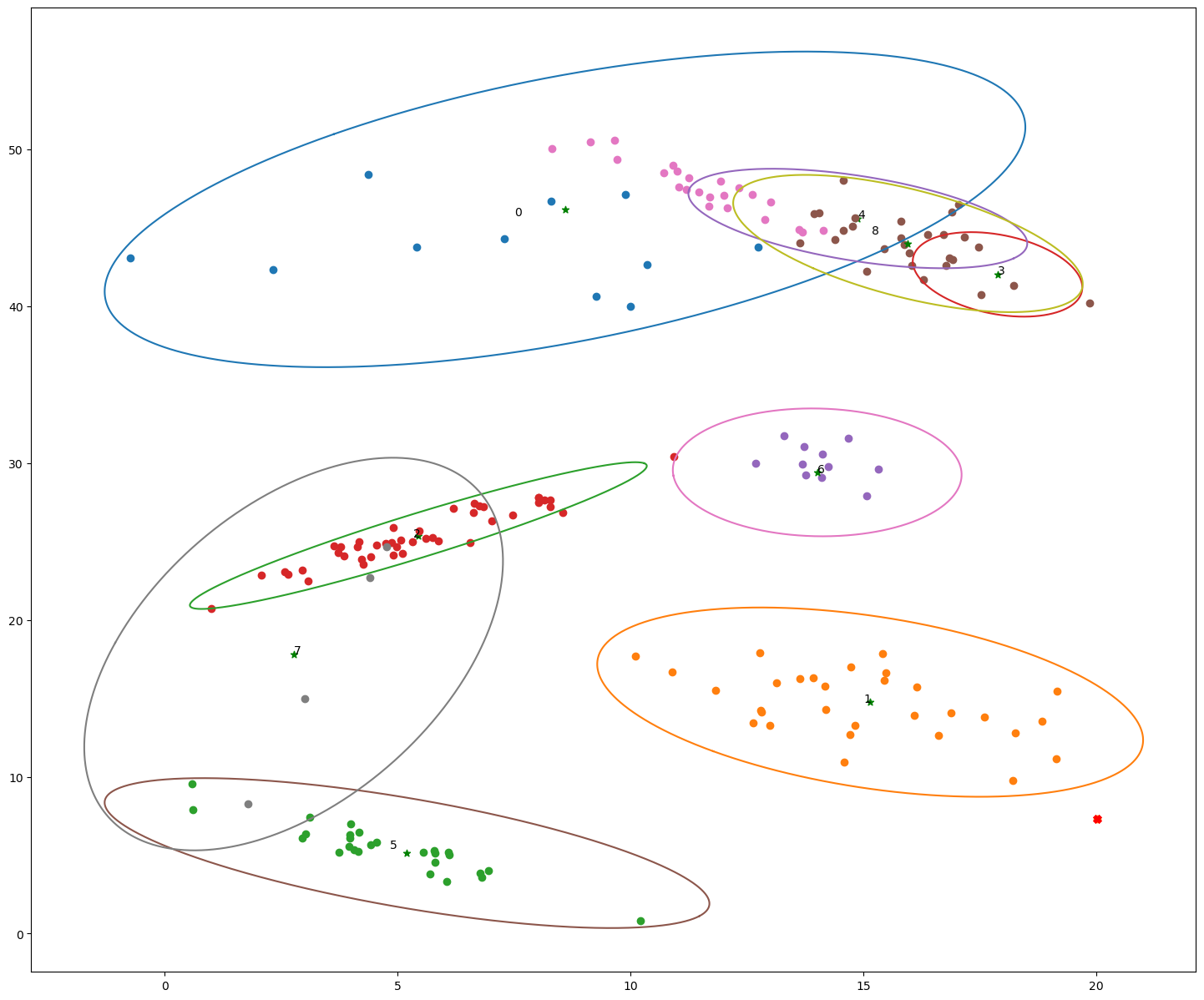}
    \caption{Signature of base clusters when k$^{th}$ data chunk arrived}
    \label{fig:basecluster}
\end{figure}

\begin{figure}[h]
    \centering
    \includegraphics[width=0.4\textwidth]{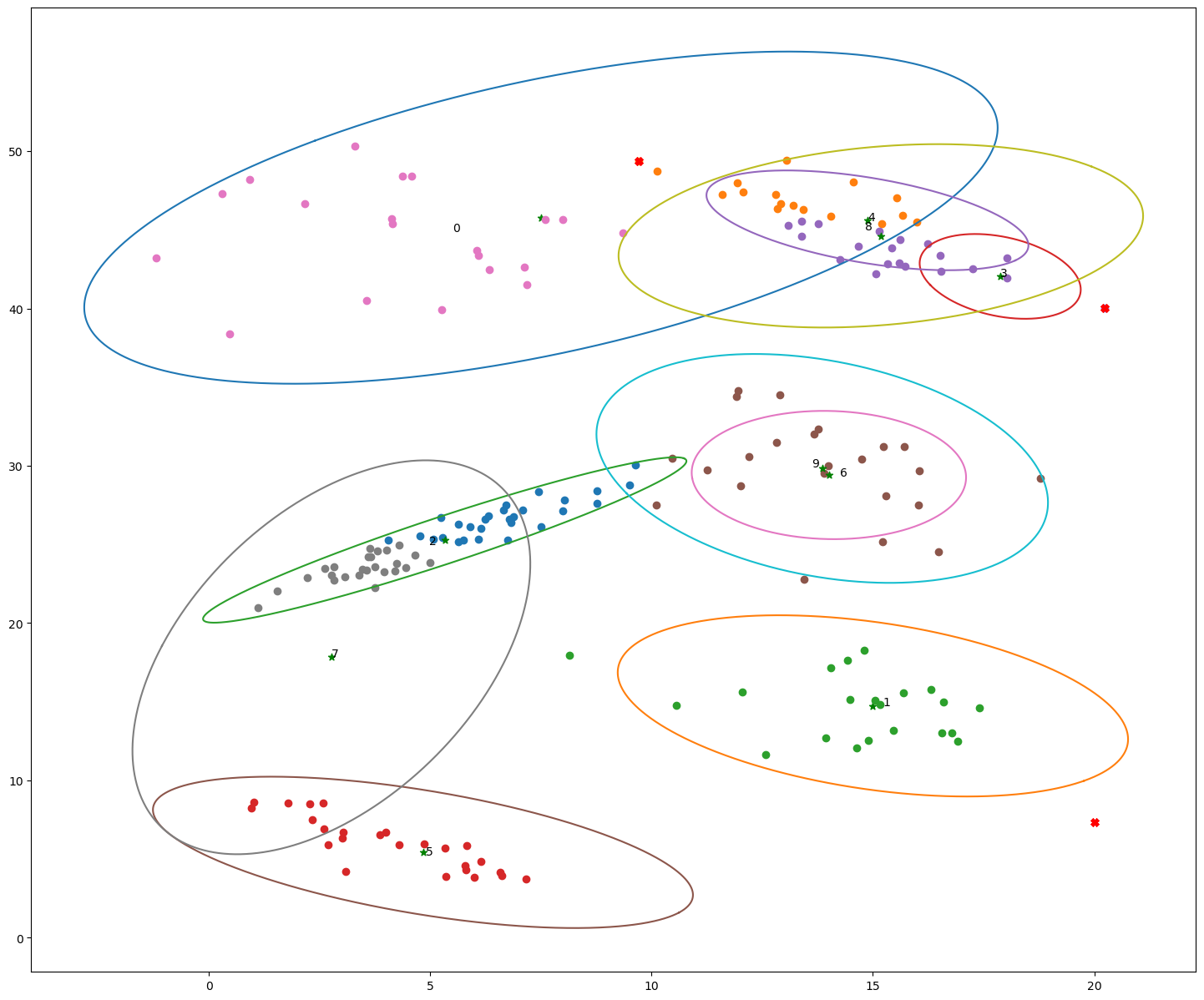}
    \caption{Signature of base clusters after k$^{th}$ data chunk merged with the base cluster}
    \label{fig:mergedclusters}
\end{figure}

We illustrate here with an example synthetic dataset the main ideas of our methodology. Figure \ref{fig:datachunk1} shows the data points for a data chunk at time $t_k$, and the results of performing GMM clustering on this chunk (each cluster is shown in different color).  Figure \ref{fig:basecluster} shows the signatures of the base set of clusters (total 9 clusters), existing in the sketch memory, summarizing the data stream preceding the current data chunk.  The ovals in this Figure represent the covariance by drawing the boundaries at Mahalanobis distance of 3.0. The red asterisks outside the ovals represent potential anomalies encountered till now in the data stream. Figure \ref{fig:mergedclusters} shows the updated signatures of the base set of clusters (total 10 clusters), after the recent chunk's results have been merged with the base set signatures. As can be seen in this figure, many of the new clusters are merged into the base set clusters while one cluster has been added to the base set. Any new anomalies detected in the recent chunk are also added to the existing set of anomalies.

\subsection{Capabilities of our proposed approach}
Our approach can provide following insights into the clusters and anomalies observed in a data stream. 
\begin{enumerate}
\item A user can enquire about clusters observed in the stream so far. If the sketch contains a large number of base clusters, a user can seek their compression into a smaller number of clusters and obtain their signatures.
\item A user can enquire about the anomalous data points of the data stream. Each such data point includes the time when it was first observed and the changes in its anomalous nature (change in its Mahalanobis distance) as the time progressed after its first observation.
\item The Mahalanobis distance of an anomalous point also changes when we compress the clusters in our sketch to fewer clusters. For each anomalous data point we can show the anomaly score profile as the clusters are compressed into fewer clusters.
\item We can show the concept drift aspects of the data stream by showing the time instants at which each of the clusters received some data in the data chunks.
\end{enumerate}

\subsection{Summary of obtained results}
We tested our methodology with two publicly available datasets, $S1$, and $unbalance$ \cite{ClusteringDatasets}. We sampled each dataset randomly, without replacement, to simulate a data stream and obtained clusters using our methodology. We then performed GMM clustering on the entire dataset simultaneously. The Rand index values for comparing the clusters obtained in these two different ways were 0.934 for the $S1$ dataset, 0.891 for the $unbalance$ dataset, and 0.908 for our synthetic dataset. These Rand index values show that our strategy of driving clustering decisions for the data stream using entropy values of clusters matches very well with the results obtained by using GMM clustering algorithm. We also produce a list of anomalous data points and temporal profiles of their mahalanobis scores.

\section{Related work}

Clustering and anomaly detection are two of the important tasks of mining data streams, and often we need to perform them simultaneously. The surveys in \cite{zubarouglu2021data,silva2013data}  summarize a number of algorithms for identifying various types of clusters, and the works in \cite{chandola2009anomaly,wang2020anomaly} review some popular approaches for anomaly detection in data streams. The clustering algorithms predominantly belong to partitional clustering paradigm \cite{liberty2016algorithm,ailon2009streaming,braverman2011streaming,chauhan2015review,de2011extending,guha2016clustering,ordonez2003clustering,wang2020anomaly,king2012online}, driven by the spirit of k-means algorithm, and some algorithms are based on density based clustering paradigm \cite{amini2014density,Chao-et-al:density-based} such as the DBSCAN algorithm. Some work also exists for identifying clusters based on Gaussian mixtures in data streams \cite{raghunathan2017learning,song2005highly,wan2018icgt}. Our approach is designed to discover Gaussian clusters in streams because these are better representation of many real-world situations. The existing work for incremental Gaussian clustering \cite{raghunathan2017learning,song2005highly,wan2018icgt} is limited to handling only diagonal covariance matrices for cluster signatures. Our proposed algorithm maintains full covariance matrices for clusters as they evolve with the new data from the stream. 

The analysis of data streams has many challenges that are not encountered when all data is available simultaneously. One such challenge is to evolve cluster signatures as new data in the stream arrives. The most common solution has been to maintain signatures of micro-clusters in some limited sketch memory and update them as the new data arrives. The CluStream algorithm presented in \cite{aggarwal2003framework}  maintains micro-clusters representing different temporal horizons and a user's query processes these microclusters in an offline manner to determine clusters for some given time horizon window. A number of algorithms have been developed that maintain a tree structure of cluster features, in the spirit of the BIRCH algorithm, and maintain and process this tree structure as new data arrives and also when queries at some clustering granularity need to be answered. Some algorithms of this type include BIRCH, E-Stream, HUE-Stream and ClusTree \cite{kranen2011clustree} \cite{ackermann2012streamkm++} \cite{meesuksabai2011hue}. Our approach is very similar to the two module approach, wherein the first module maintains and updates cluster signatures and the second module summarizes the signatures to answer user queries. In our sketch memory, we maintain a full covariance matrices and other information for a large number of clusters without structuring them in a tree hierarchy.

For anomaly detection in simultaneous processing of all data a number of methods requiring elaborate computations are popular. Local outlier factor \cite{breunig2000lof} and many of its derivatives \cite{pokrajac2007incremental,salehi2016fast} need almost all data or make assumptions such as all clusters being spherical in nature. In our approach we define anomalous nature of a point i terms of its Mahalanobis distance from its nearest cluster and keep updating these scores as the cluster signatures evolve.

\section{Main steps of our approach}
Our objectives include identifying Gaussian clusters in the stream data, and also identifying those data points that are likely to be outliers from the perspective of these clusters. Due to the streaming nature of the data the characteristics of Gaussian clusters evolve and are updated after arrival of each new data chunk. The anomalous nature of each suspected outlier is also updated at each time point in the context of the evolving cluster characteristics. 

To perform these tasks we maintain a Sketch memory for the data stream in which summary information about the clusters and anomalous data points are maintained. After each data chunk is processed these summaries in the Sketch are updated. 
 
\subsection{Sketch memory contents}
We need to store in the sketch details for each evolving cluster. We also need to store in the sketch information about each potential anomalous data point encountered in the stream so far. We assume that at each time step a new data chunk arrives from the stream. We use the Gaussiam Mixture Model (GMM) algorithm on the data in this chunk to form clusters. The clusters found in the very first data chunk, are used to create signatures for the base set of clusters stored in the sketch memory. The data points of the chunk are then discarded. The sketch contains the following information as signature for each of the clusters: (i) total number of data points in the cluster, (ii) covariance matrix of the cluster, and (iii) centroid of the cluster. 

We also look for anomalous data points  in each new data chunk and record them in the sketch. This is done by computing the Mahalanobis distance of each data point of the chunk from the clusters identified in the chunk. All those points whose Mahalanobis distance values are above some pre-set threshold are saved in the sketch as potential anomalies.

\begin{algorithm}
\caption{Temporal module clustering and Anomaly Detection}\label{algo1}
\begin{algorithmic}[1]
\Require \\
S={X1, X2, X3, …, Xn}; where S= Data stream, \\
Xi = Data chunk at time i, \\
Ti={t1,t2,t3, …, tn} \\
K=number of cluster (overestimated number chosen) 
\Ensure  Temporal\_Signature
\State $GMM \Rightarrow X1$
\State $Base\_signature \Rightarrow {C1, C2, C3, …, Cn}$
\State At time t process $Xi$ \& create Cluster\_signature(Xi)
\State Find four nearest cluster signature for every cluster in Cluster\_signature(Xi)
\For{$Each four nearest cluster Base_signature(i)$}                    
        \State Calculate e1, e2 and $\Delta$e1, $\Delta$e2
        \If{$\Delta$ e1 $<0$ \& $\Delta$ e2 $<5$}\label{algln2}
        \State $Merge Cluster\_signature(Xi) with Base\_signature$
        \Else
                \State $Add Cluster\_signature(Xi) into Base\_signature$
        \EndIf
\EndFor
\For{Every data point in Xi}
    \State Calculate Mahalanobis distance from the center of the cluster
    \If{$Mahalanobis_Distance > 3$}
        \State Save it as anomaly
    \EndIf
    \State Delete raw data point
\EndFor
\State $Base\_signature \Rightarrow Temporal\_Signature$
\end{algorithmic}
\end{algorithm}

\subsection{Updates to sketch memory for new data chunks}
When the next data chunk arrives we perform the GMM algorithm on the data of this new chunk, find the cluster signatures of this chunk and use them to update the cluster signtaures in our base set. It may result in modifications to the signatures of the current base set signatures, and some new cluster signatures may be added to the base set. The algorithmic details of the update process are included in a following section. The Mahalanobis distances for the anomalous points stored in the sketch are recomputed and recorded in light of the updated cluster signatures, and some new anomalous points, identified in the new data chunk, may be added to the sketch.

At a time instant $t$, when a new data chunk arrives, clusters are formed for this chunk using the GMM algorithm. This clustering algorithm takes the number of clusters as a parameter and we generally choose to form much larger number of clusters than what a user may want to see in a summary for the data stream. At any time when a user wants to query the sketch to get the results we merge the larger number of small clusters in the base set to form either the desired number of clusters or the optimal number of clusters. This compression of smaller sized clusters is performed by the {\it Compression Module} of our approach. 

\subsubsection{Merging of Cluster Signatures}
\label{ClusterMergingSection}

Let us say our base set of clusters in the sketch contains clusters $B_1, B_2, . . . B_m$. For each $i^{th}$ cluster the sketch contains the quantities $B\_Mean_i$, $B\_Cov_i$, and $B\_NumPoints_i$. Similarly, for the new data chunk we now have the clusters $C_1, C_2, . . .C_k$, and corresponding quantities $C\_Mean_j$, $C\_Cov_j$, and $C\_NumPoints_j$. 

Our goal now is to find for each chunk cluster whether there is a base cluster that is a good candidate for absorbing it, and if there is one, then merging the two to form a new cluster signature. The decisions to merge are driven by the following entropy based intuitive notion and criteria. A closely packed cluster has low entropy and a widely spread out cluster has large entropy. We can compute the entropy of a single cluster $B_i$, using its covariance matrix, as:
\begin{equation}
entropy(B_i) = n/2 * log_2((2 \pi e) +log_2 \begin{vmatrix} B\_Cov_i \end{vmatrix})
\end{equation}

where $n$ is the number of dimensions of the data, and $\begin{vmatrix} B\_Cov_i \end{vmatrix}$ is the determinant of the covariance matrixs. When a base cluster $B_i$ and a chunk cluster $C_j$ are considered for merger, we need to construct the signature of the  potential merged cluster $P$. This signature for the potential cluster $P$ can be computed as:
\begin{equation}
P\_NumPoints = B\_NumPoints_i + C\_NumPoints_j
\end{equation}
\begin{equation}
\begin{split}
P\_mean = (B\_Mean_i * B\_NumPoints_i) + \\ 
    (C\_Mean_j * C\_NumPoints_j)
\end{split}
\end{equation}
\begin{equation}
\begin{aligned}
P\_Cov = ((B\_NumPoints_i*B\_Cov_i+ \\
C\_NumPoints_j*C\_Cov_j) + \\
B\_NumPoints_i*(B\_Mean_i-P\_Mean)^T\\ 
*(B\_Mean_i-P\_Mean) + \\
C\_NumPoints_j*(C\_Mean_j-P\_Mean)^T*\\
(C\_Mean_j-P\_Mean)) / P\_NumPoints
\end{aligned}
\end{equation}

From this signature the entropy of the potential merged cluster $P$ can be computed as:
\begin{equation}
entropy(P) = n/2 * log_2((2 \pi e) +log_2 \begin{vmatrix} P\_Cov \end{vmatrix})
\end{equation}

From an intuitive perspective a base cluster and a chunk cluster are good candidates for merger if the entropy of their merged cluster is smaller than the entropy of each of them, or if there is an increase in entropy after merger, and it is very small. This intuitive criterion will ensure that the entropy of the entire clustering remains low. 

We implement the above intuitive condition for merger as follows. 
\begin{itemize}
    \item e1 = entropy of candidate base cluster $B_i$
    \item e2 = entropy of the candidate chunk cluster $C_j$
    \item e = entropy of cluster formed by merging $B_i$ and $C_j$
    \item $\Delta$e1 = ((e - e1)/e1)*100
    \item $\Delta$e2 = ((e - e2)/e2)*100
\end{itemize}
 One can choose some thresholds for  $\Delta$e1 and $\Delta$e2 to decide whether to merge these two clusters. In our implementation we choose to merge the clusters as follows:
 \begin{itemize}
     \item If the chunk cluster has ten or more data points then merge the clusters if $\Delta$e1 is less than 5\% and $\Delta$e2 is less than 10\%
 \end{itemize}
 \begin{itemize}
     \item If the chunk cluster has fewer than ten data points then merge the clusters if $\Delta$e1 is less than 10\% 
 \end{itemize} 
 
The percentage thresholds can affect only the number of clusters in the base set, and the compression module will later reduce them to fewer clusters no matter how many are there in the base set. The reason for using two different thresholds is that the base clusters are assumed to be larger in size and more stable and therefore less likely to change with the merger of a new smaller cluster from the chunk. A larger increase in the entropy of the chunk cluster is acceptable because of it containing fewer data points. 


It is possible that a chunk cluster may meet the above criteria for merger with more than one base cluster. To determine the best candidate for merger among these possible candidate base clusters we perform the following steps:
\begin{enumerate}
\item For each chunk cluster find four nearest base clusters in terms of Euclidean distances between their centroids
\item Form a potential merged cluster of the chunk cluster with each of these four base clusters.
\item Compute the entropy of each potential merged cluster, and their $\Delta$e1 and $\Delta$e2 values.
\item Find the pair with smallest $\Delta$e1, if there is a tie then prefer the one with smaller $\Delta$e2.
\item If there is still a tie then pick any one pair randomly.
\end{enumerate}

As a result of the above process some chunk clusters may get merged with some base clusters and some chunk clusters may not find any base set candidates for merger. All the unmerged chunk clusters are added to the base set as independent clusters. 
The number of clusters in the base set, thus, can keep increasing. The compression module of our approach processes the clusters in the base set to look for possible mergers so that the number of clusters in the base set can be reduced.

\subsubsection{Compression Module} 
The compression module seeks to merge those clusters of the base set whose merger does not cause significant increase in the entropy of the clusters. This results in the reduction in the total number of clusters in the base set. Similar to the process of candidate selection for merger during the temporal evolution of clustering, as shown in section \ref{ClusterMergingSection}, during compression we consider each cluster in the base set and find its four nearest other clusters using the Euclidean distance between their centroids.   

We employ the same quantities, $\Delta$e1 and $\Delta$e2 as in Section \ref{ClusterMergingSection}, and can use different thresholds for merger decisions during this compression process. If we decide to merge two clusters when $\Delta$e1 and $\Delta$e2 are both less than zero, then our resulting merged clusters are those situations in which the entropy of the merged cluster is lower than the original entropy of each individual cluster.

We can relax the merger criterion by selecting higher values for $\Delta$e1 and $\Delta$e2. This will allow clusters to be merged with some permissible levels of increase in entropy, and therefore resulting in smaller number of final clusters.
\begin{figure}[!ht]
    \centering
    \includegraphics[width=0.5\textwidth]{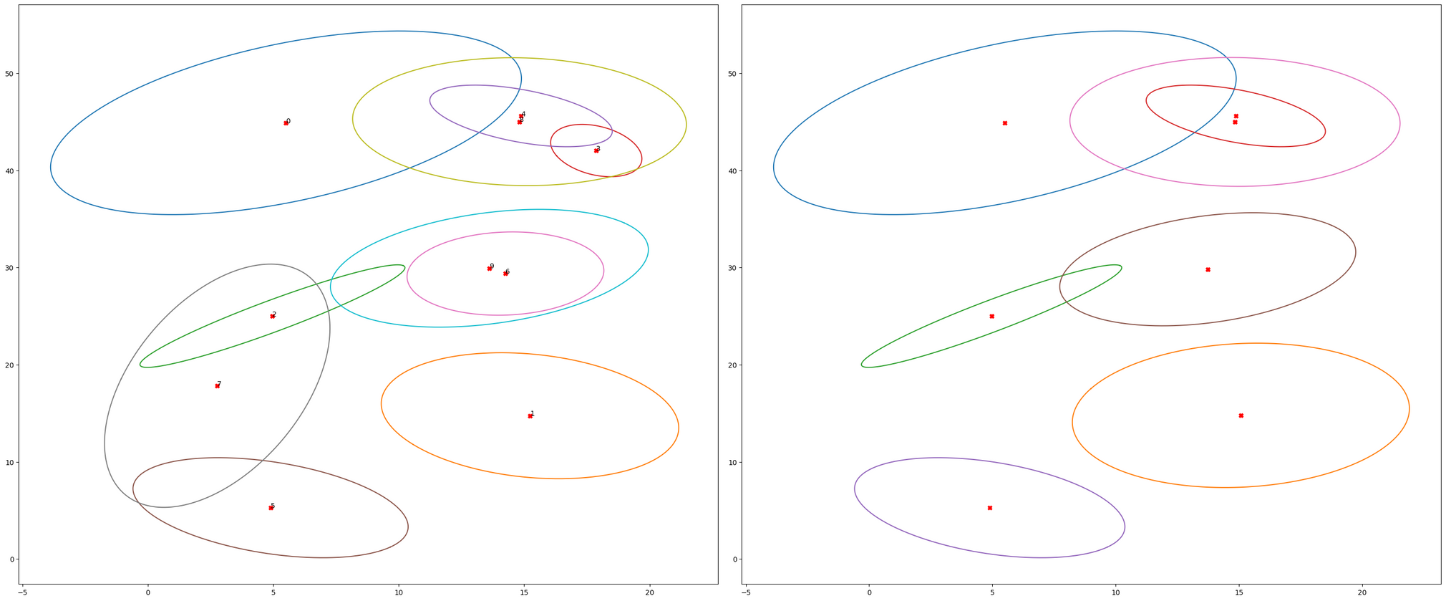}
    \caption{Synthetic data with 10 clusters (left) and 7 clusters (right)}
    \label{fig:Synthetic-compression}
\end{figure}

Figure \ref{fig:Synthetic-compression} shows the results of applying the compression module to the ten clusters of the base set (left frame) for our synthetic data, and the right frame shows the seven resulting clusters. This compression module can be executed whenever the number of clusters in the base set may become too large, or at the time of query by a user requiring clusters with some specified granularity.

\subsection{Management of Anomalies}
When a new data chunk arrives and GMM clustering is performed on its data, we search for anomalous data points as follows:
\begin{enumerate}
    \item If the new chunk cluster is merged with a base cluster then compute the Mahalanobis distance of each data point of the chunk cluster from the merged cluster.
    \item If the new chunk cluster is not merged with a base cluster, and is added as a new member of the base set, then compute the Mahalanobis distance for each data point of the chunk cluster from its own cluster.
    \item All those data points of the new chunk whose Mahalanobis distance is above some threshold (say, 3.0, for example), are considered as anomalous and are added to the sketch memory.
\end{enumerate}
Mahalanobis distance has been used extensively for determining the anomalous nature of data points in Gaussian environments \cite{ghorbani2019mahalanobis}.

\subsection{Evolution of anomaly scores}
When a data point is first detected as an anomaly, it is stored in the sketch memory along with its coordinates, identity of its associated cluster, and its Mahalanobis score. At every subsequent time instant when a new data chunk arrives, the signatures of clusters in the base set may change due to some cluster mergers. Also, each new data chunk may cause some new clusters to be added to the base set. This evolution of the base set of clusters affects the Mahalanobis score of each recorded anomalous data point. At the end of processing each data chunk, we, therefore, revise the Mahalanobis distance of each anomalous point recorded in our sketch memory. This re-evaluation may cause an anomalous data point's Mahalanobis distance to increase or decrease, and also there may be a change in the identity of the cluster nearest to an anomalous data point.

\subsection{Temporal and compression profile of detected anomalies.}
Our sketch memory maintains a temporal record of all Mahalanobis distance values of each anomalous data point. This helps us with the insight about changes in anomalous nature of each data point with time, as more data arrives in the stream. Figure \ref{fig:anomalies_temporal_profile} depicts the change in the anomaly score as their temporal profiles. We can observe that anomaly score for index-0, index-78, index-214 decreases significantly as more data chunks arrive. Conversely, anomaly scores for index-13, index-57, and index-136 increase or remain same with the arrival of new data chunks. 


The compression module merges many of the clusters in the base set, resulting in fewer clusters remaining in the sketch memory. Since the anomaly score of each anomalous data point is determined as its minimum Mahalanobis distance from one of its neighboring clusters, any changes in clustering due to mergers of base set clusters can potentially affect the anomaly scores of the anomalous data points in the sketch.

We capture the behaviour of the anomaly score of each anomalous data point as the number of clusters decreases due to compression. Every time two base clusters are merged we recompute the anomaly score of each anomalous data point and record it. 

At the end of the compression module we can see the profile of anomaly scores of each anomalous data point as the number of clusters decreases. As shown in Figure \ref{fig:anomalies_compression_profile} we can see that for anomalies index-174, index-73, index-119 Mahalanobis distance increases even when fewer clusters remain after compression. Whereas for anomaly index-24, index-45, index-3 and index-147 Mahalanobis distance decreses significantly with fewer clusters and these points are no longer anomalous.

\subsection{Concept Drift Detection}

\begin{figure}
    \centering
    \includegraphics[width=0.5\textwidth]{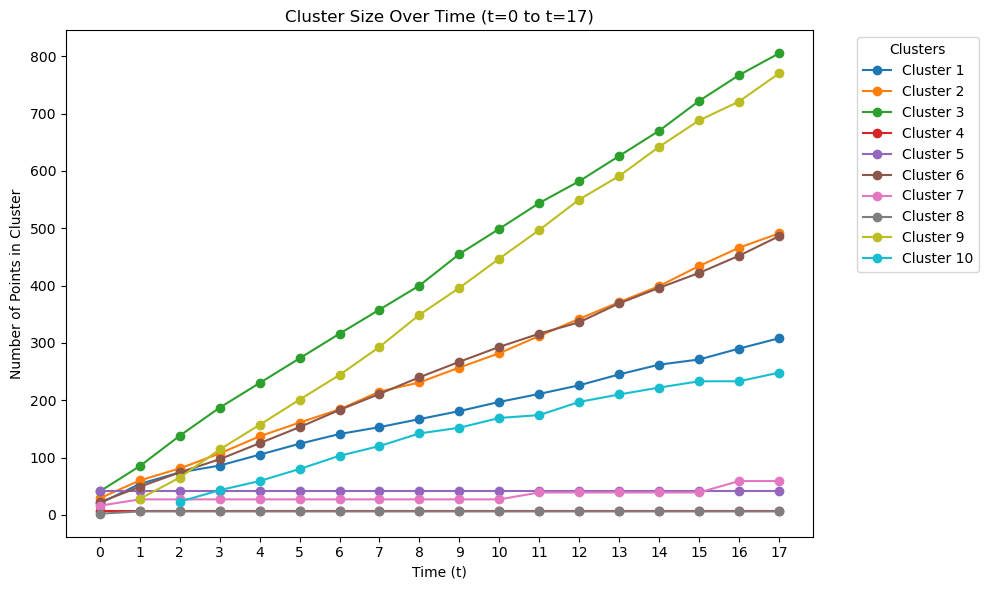}
    \caption{Concept drift in Synthetic dataset}
    \label{fig:concept-drift-synthetic}
\end{figure}

As clusters from new data chunks are merged with the clusters in the base set, we modify the signatures of the evolving clusters. We also record the time points at which new data was merged into a cluster in the base set. An examination of the time stamps at which new data was added to an evolving cluster provides a good insight into the concept drift taking place in the data stream. Figure \ref{fig:concept-drift-synthetic} shows the growth of the clusters in the synthetic dataset and the new emerging clusters (cluster 9 starting at time index 1 and cluster 10 starting at time index 2) as a new data chunk arrives at those times. 
Lot of work is needed to handle concept drift when we are constructing a supervised learning system,\cite{gama2014survey} or are interested in clusterings for different time windows in the past.\cite{aggarwal2003framework} In the work presented here our focus is on obtaining Gaussian clusters incrementally and only detecting the nature of concept drift during the arrival of data.

\section{Results}

Our methodology described above has been verified using a syntheic dataset and two publicly available benchmark datasets $S1$, and $Unbalance$ \cite{ClusteringDatasets}. All these are sets of two-dimensional data points. Each data set is augmented with by appending to it its own two more copies to increase the overall data size, without affecting the essential nature of clusters present in each of them. We then randomly select uniform sized data chunks from the augmented datasets to simulate a data stream.

\subsection{Validation of Our Clustering Results}
Using our methodology we generated Gaussian clusters from the data streams simulated from the augmented datasets. For the benchmark cases we clustered all data points of each dataset using GMM algorithm. We then computed the RAND index for the clustering obtained for the same datasets in two different ways.

\subsection{Results with '$S1$' dataset}
\begin{figure}[!ht]
    \centering
    \includegraphics[width=0.5\textwidth]{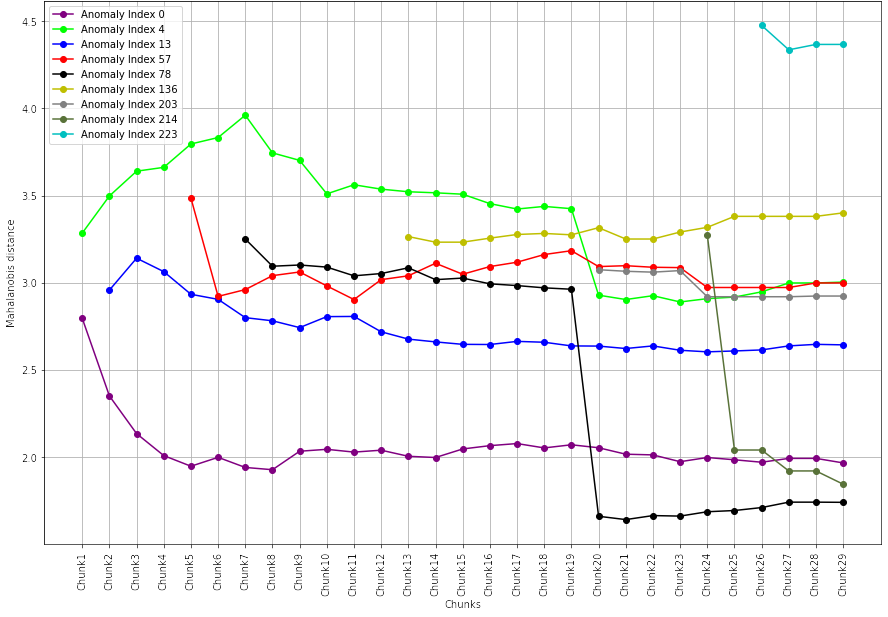}
    \caption{Temporal profile of anomalies in $S1$ dataset}
    \label{fig:anomalies_temporal_profile}

    \centering
    \includegraphics[width=0.5\textwidth]{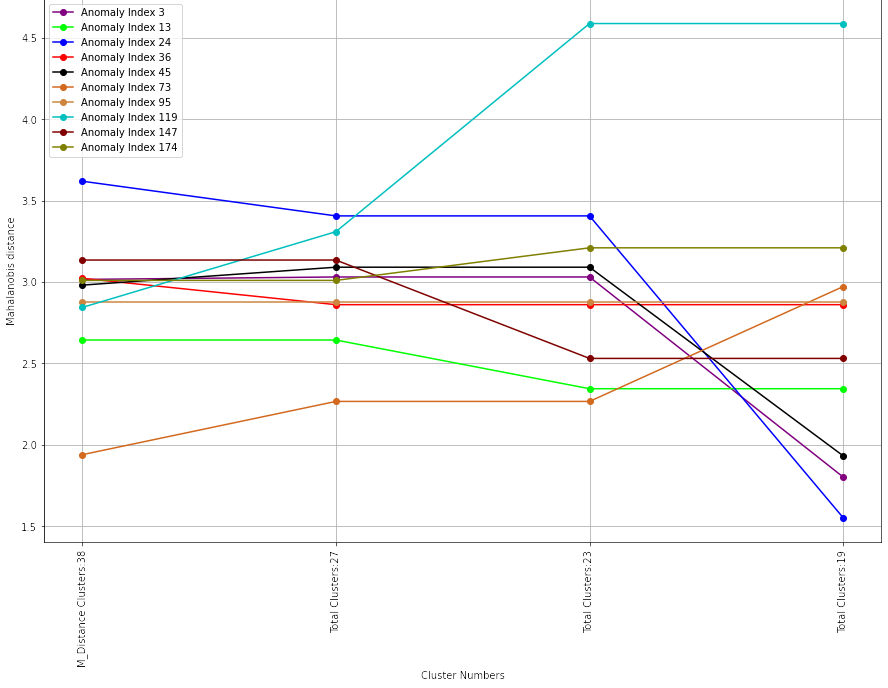}
    \caption{Compression profile of anomalies in $S1$ dataset}
    \label{fig:anomalies_compression_profile}

\end{figure}

\begin{figure}[!ht]
    \centering
    \includegraphics[width=0.5\textwidth]{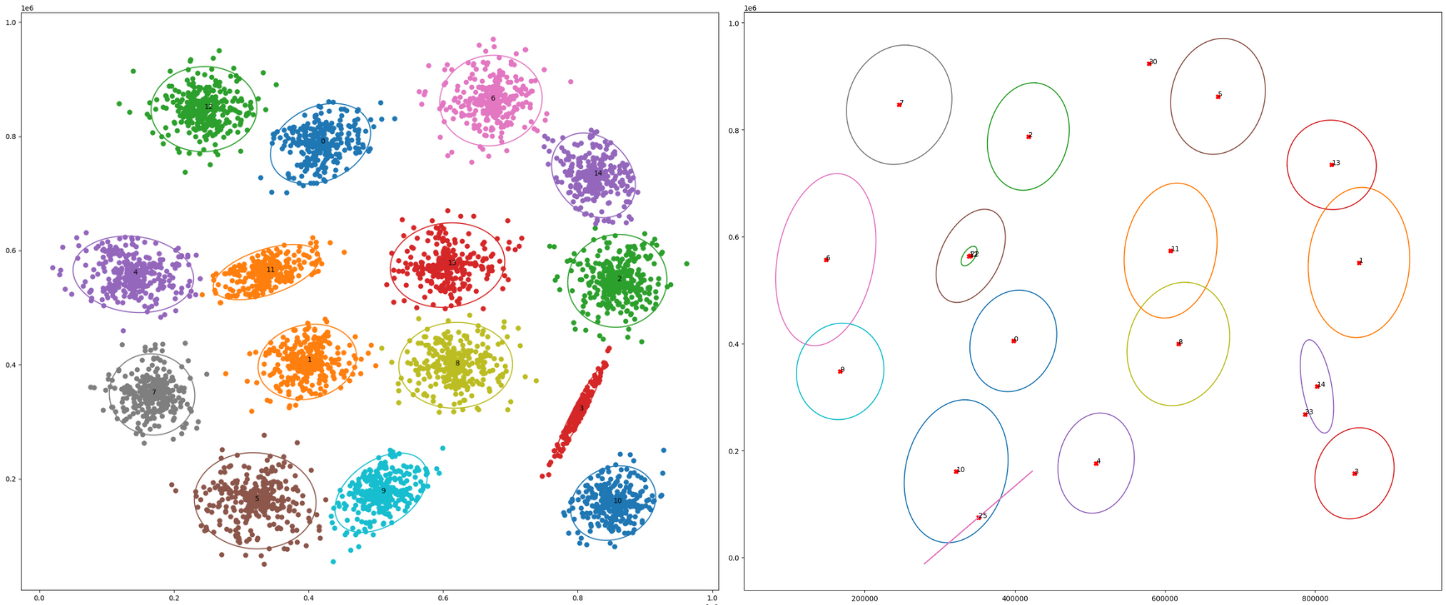}
    \caption{$S1$ dataset benchmark result (left) vs our approach (right)}
    \label{fig:S1result}
\end{figure}

The augmented version of this dataset has 15000 points and it contains 15 clusters as ground truth. 
We chose to create 30 Gaussian clusters in each data chunk of 500 data points and then used them to update the base set signatures as appropriate. Anomalies are detected in each chunk after clustering the data points of the chunk. The data points having Mahalanobis distance greater than 3.0 are termed as anomalies and stored in the sketch memory along with the base set cluster signatures. For dataset $S1$ Figure \ref{fig:S1result} shows the final clustering results obtained by our streaming approach, and also for the benchmark case. We can see that the clusters found in both cases are very similar in characteristics. Rand Index for their comparison turns out to be 0.934 which shows a very good match between two clusterings. Many anomalous points were identified during the processing of data stream from $S1$ and their evolving Mahalanobis scores were recorded as cluster shapes evolved. Figure \ref{fig:anomalies_temporal_profile} shows the changing anomaly scores for the detected anomalies in temporal profile. When the number of clusters in the base set was reduced using our compression module, the anomalies again changed their scores due to the changes in the signatures of the clusters. This pattern of evolving anomaly scores, as the number of clusters reduces gradually, is shown in Figure \ref{fig:anomalies_compression_profile}. 

\begin{figure}[!ht]
    \centering
    \includegraphics[width=0.5\textwidth]{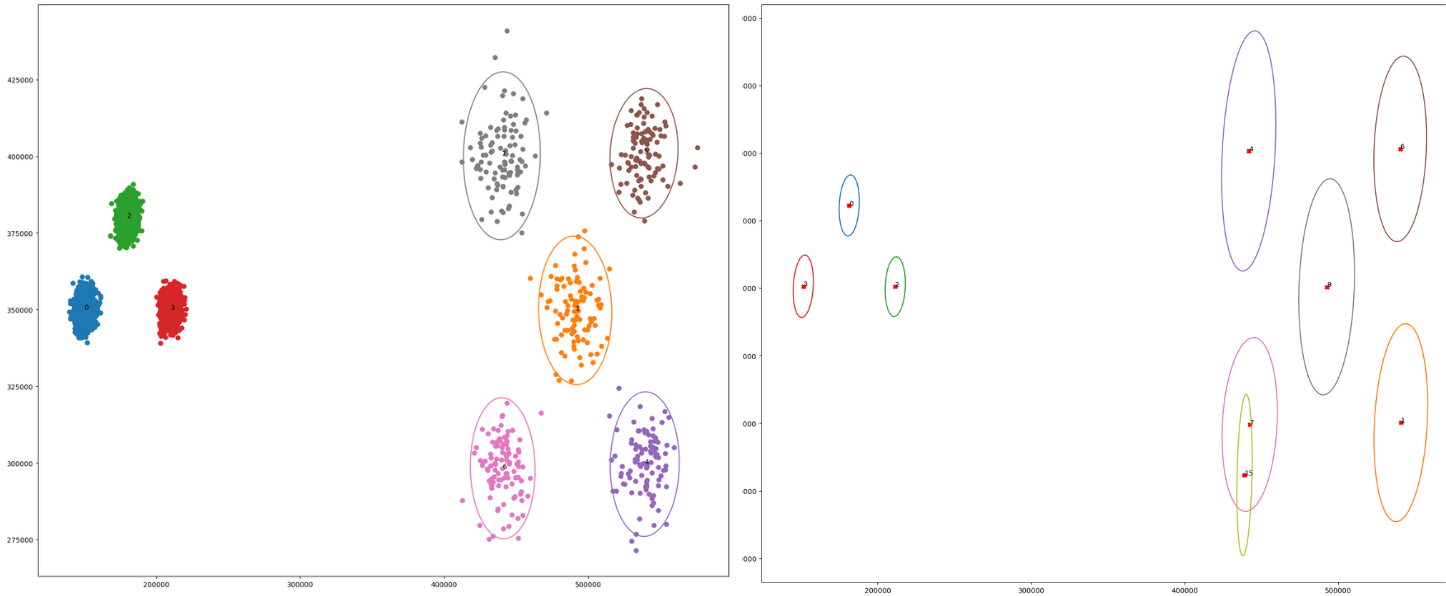}
    \caption{$unbalance$ dataset benchmark result (left) vs our approach (right)}
    \label{fig:unblance-result}
\end{figure}

\subsection{Results with '$Unbalance$' dataset}
\begin{figure}[h]
    \centering
    \includegraphics[width=0.4\textwidth]{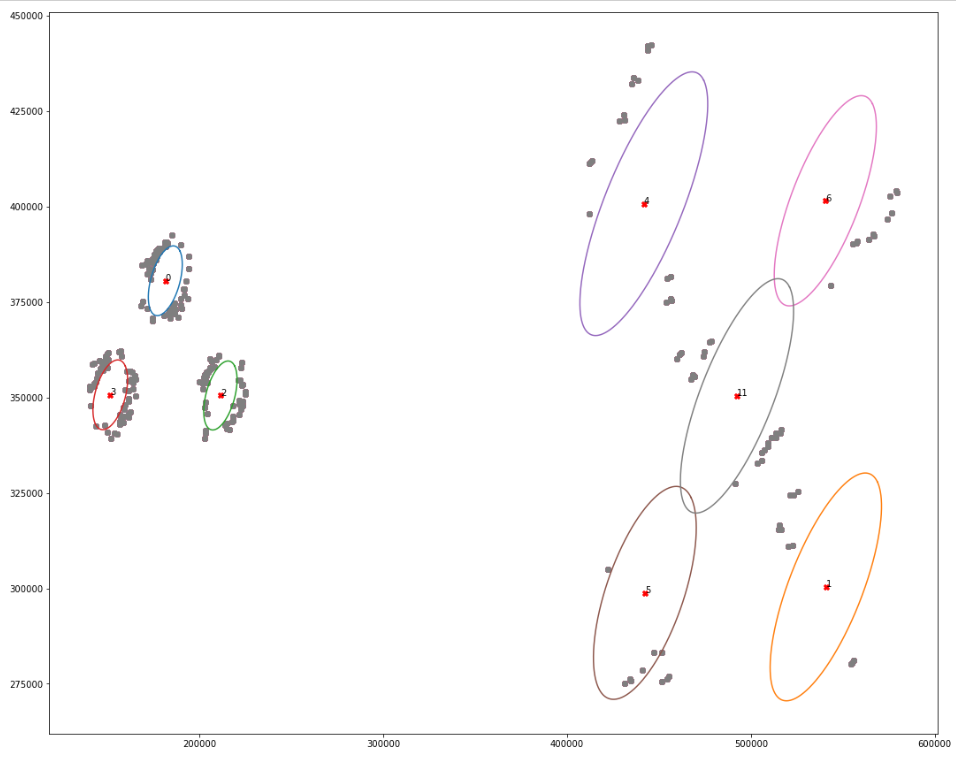}
    \caption{Anomalies location in unbalnced dataset with 8 clusters}
    \label{fig:anomalies-unbalanced}
\end{figure}

This dataset has 6500 datapoints, resulting in 19500 points in the augmented dataset. We created data chunks of randomly selected 500 points each to simulate the data stream. Tests for the $Unbalance$ dataset show the clustering results for the benchmark case for our approach in Figure \ref{fig:unblance-result}. The Rand Index value for this case is 0.891. Figure \ref{fig:anomalies-unbalanced} shows the anomalies detected after temporal and compression profile when there are 8 clusters in the base cluster signature. 

\subsection{Interpretation of results and contribution}
Results obtained by following our two module approach helps us find key insights like Gaussian clusters and locations and observation times of anomalies in the data stream. Our results show that using an entropy based approach we obtain almost the same clusters from the stream that are obtained by a GMM algorithms working on all the data simultaneously. Rand index values close to 0.9 show that our clusters are very close to those in the original data and there has been very little loss of information due to the streaming effect.

Our approach has been demonstrated with 2-dimensional datasets, primarily for good visualization of results, but it is easily generalizable to higher dimensional datasets without requiring any changes to any of the steps of our methodology.

Also, we can obtain deep insights into the detected anomalies. Figure \ref{fig:anomalies_temporal_profile} shows that the anomaly scores of some detected anomalous points change, sometimes little and sometimes significantly, as the newly arriving data change the signatures of clusters near these anomalous data points.

We are also able to capture the concept drift taking place in the data stream. We record the time points and the number of new data points that modify the signature of each base cluster. We can visualize this concept drift information as a temporal profile for a cluster's growth, as the contributions to clusters at each time point, or as the population growth trajectory of each cluster. Figure \ref{fig:concept-drift-synthetic} shows the last type of visualization for the synthetic dataset. We can obtain concept drift plots for other datasets also.

\subsection{Limitation of approach}
Our algorithm uses some parameters such as the number of clusters in the base set, entropy change thresholds for merging clusters from new data chunks into base set clusters, and the Mahalanobis distance threshold for marking points as anomalous. The clustering related thresholds do not affect the results much because the compression module gradually relaxes the merger criteria to compress the base set clusters into desired smaller number of clusters. For anomaly detection we mark a data point as anomalous only when it arrives as part of a data chunk. Its anomaly score is updated as cluster signatures change. Some data points which may not be anomalous when they arrive but may be anomalous in the larger context of the data yet to arrive may be lost. We can keep a lower threshold for Mahalanobis distance to store more points as potential anomalies to not lose some of those points that may later emerge as more prominent anomalies.

\section{Conclusion}
In this paper we have presented a methodology for obtaining Gaussian clusters and anomalies from a streaming data. This approach can maintain and update complete covariance matrix for each component cluster as the new data chunks arrive. The anomalies are identified and their anomaly scores are updated as new data arrives. We also capture and illustrate concept drift in data in terms of the rates at which various clusters receive their data along time. We have illustrated successful working of this approach with one synthetic and two public datasets. These contributions improve the capabilities of many other existing algorithms for clustering and anomaly detection with streaming data.

\bibliographystyle{IEEEtran}
\bibliography{IncrLearn}

\begin{thebibliography}{10}
\providecommand{\url}[1]{#1}
\csname url@samestyle\endcsname
\providecommand{\newblock}{\relax}
\providecommand{\bibinfo}[2]{#2}
\providecommand{\BIBentrySTDinterwordspacing}{\spaceskip=0pt\relax}
\providecommand{\BIBentryALTinterwordstretchfactor}{4}
\providecommand{\BIBentryALTinterwordspacing}{\spaceskip=\fontdimen2\font plus
\BIBentryALTinterwordstretchfactor\fontdimen3\font minus \fontdimen4\font\relax}
\providecommand{\BIBforeignlanguage}[2]{{%
\expandafter\ifx\csname l@#1\endcsname\relax
\typeout{** WARNING: IEEEtran.bst: No hyphenation pattern has been}%
\typeout{** loaded for the language `#1'. Using the pattern for}%
\typeout{** the default language instead.}%
\else
\language=\csname l@#1\endcsname
\fi
#2}}
\providecommand{\BIBdecl}{\relax}
\BIBdecl

\bibitem{zubarouglu2021data}
A.~Zubaro{\u{g}}lu and V.~Atalay, ``Data stream clustering: a review,'' \emph{Artificial Intelligence Review}, vol.~54, no.~2, pp. 1201--1236, 2021.

\bibitem{silva2013data}
J.~A. Silva, E.~R. Faria, R.~C. Barros, E.~R. Hruschka, A.~C.~d. Carvalho, and J.~Gama, ``Data stream clustering: A survey,'' \emph{ACM Computing Surveys (CSUR)}, vol.~46, no.~1, pp. 1--31, 2013.

\bibitem{chandola2009anomaly}
V.~Chandola, A.~Banerjee, and V.~Kumar, ``Anomaly detection: A survey,'' \emph{ACM computing surveys (CSUR)}, vol.~41, no.~3, pp. 1--58, 2009.

\bibitem{wang2020anomaly}
Z.~Wang, Y.~Zhou, and G.~Li, ``Anomaly detection by using streaming k-means and batch k-means,'' in \emph{2020 5th IEEE international conference on big data analytics (ICBDA)}.\hskip 1em plus 0.5em minus 0.4em\relax IEEE, 2020, pp. 11--17.

\bibitem{ailon2009streaming}
N.~Ailon, R.~Jaiswal, and C.~Monteleoni, ``Streaming k-means approximation,'' \emph{Advances in neural information processing systems}, vol.~22, 2009.

\bibitem{braverman2011streaming}
V.~Braverman, A.~Meyerson, R.~Ostrovsky, A.~Roytman, M.~Shindler, and B.~Tagiku, ``Streaming k-means on well-clusterable data,'' in \emph{Proceedings of the twenty-second annual ACM-SIAM symposium on Discrete Algorithms}.\hskip 1em plus 0.5em minus 0.4em\relax SIAM, 2011, pp. 26--40.

\bibitem{chauhan2015review}
P.~Chauhan and M.~Shukla, ``A review on outlier detection techniques on data stream by using different approaches of k-means algorithm,'' in \emph{2015 international conference on advances in computer engineering and applications}.\hskip 1em plus 0.5em minus 0.4em\relax IEEE, 2015, pp. 580--585.

\bibitem{de2011extending}
J.~de~Andrade~Silva and E.~R. Hruschka, ``Extending k-means-based algorithms for evolving data streams with variable number of clusters,'' in \emph{2011 10th International Conference on Machine Learning and Applications and Workshops}, vol.~2.\hskip 1em plus 0.5em minus 0.4em\relax IEEE, 2011, pp. 14--19.

\bibitem{guha2016clustering}
S.~Guha and N.~Mishra, ``Clustering data streams,'' in \emph{Data stream management: processing high-speed data streams}.\hskip 1em plus 0.5em minus 0.4em\relax Springer, 2016, pp. 169--187.

\bibitem{ordonez2003clustering}
C.~Ordonez, ``Clustering binary data streams with k-means,'' in \emph{Proceedings of the 8th ACM SIGMOD workshop on Research issues in data mining and knowledge discovery}, 2003, pp. 12--19.

\bibitem{amini2014density}
A.~Amini, T.~Y. Wah, and H.~Saboohi, ``On density-based data streams clustering algorithms: A survey,'' \emph{Journal of Computer Science and Technology}, vol.~29, pp. 116--141, 2014.

\bibitem{raghunathan2017learning}
A.~Raghunathan, P.~Jain, and R.~Krishnawamy, ``Learning mixture of gaussians with streaming data,'' \emph{Advances in Neural Information Processing Systems}, vol.~30, 2017.

\bibitem{song2005highly}
M.~Song and H.~Wang, ``Highly efficient incremental estimation of gaussian mixture models for online data stream clustering,'' in \emph{Intelligent Computing: Theory and Applications III}, vol. 5803.\hskip 1em plus 0.5em minus 0.4em\relax SPIE, 2005, pp. 174--183.

\bibitem{wan2018icgt}
Y.~Wan, X.~Liu, Y.~Wu, L.~Guo, Q.~Chen, and M.~Wang, ``Icgt: A novel incremental clustering approach based on gmm tree,'' \emph{Data \& Knowledge Engineering}, vol. 117, pp. 71--86, 2018.

\bibitem{aggarwal2003framework}
C.~C. Aggarwal, S.~Y. Philip, J.~Han, and J.~Wang, ``A framework for clustering evolving data streams,'' in \emph{Proceedings 2003 VLDB conference}.\hskip 1em plus 0.5em minus 0.4em\relax Elsevier, 2003, pp. 81--92.

\bibitem{kranen2011clustree}
P.~Kranen, I.~Assent, C.~Baldauf, and T.~Seidl, ``The clustree: indexing micro-clusters for anytime stream mining,'' \emph{Knowledge and information systems}, vol.~29, pp. 249--272, 2011.

\bibitem{ClusteringDatasets}
\BIBentryALTinterwordspacing
P.~Fr\"anti and S.~Sieranoja, ``K-means properties on six clustering benchmark datasets,'' pp. 4743--4759, 2018. [Online]. Available: \url{http://cs.uef.fi/sipu/datasets/}
\BIBentrySTDinterwordspacing

\bibitem{liberty2016algorithm}
E.~Liberty, R.~Sriharsha, and M.~Sviridenko, ``An algorithm for online k-means clustering,'' in \emph{2016 Proceedings of the eighteenth workshop on algorithm engineering and experiments (ALENEX)}.\hskip 1em plus 0.5em minus 0.4em\relax SIAM, 2016, pp. 81--89.

\bibitem{king2012online}
A.~King, ``Online k-means clustering of nonstationary data,'' \emph{Prediction Project Report}, pp. 1--9, 2012.

\bibitem{Chao-et-al:density-based}
F.~Cao, M.~E.~W. Qian, and A.~Zhou, ``Density-based clustering over an evolving data stream with noise.''\hskip 1em plus 0.5em minus 0.4em\relax Bethesda, MD, USA: Proceedings of the Sixth SIAM International Conference on Data Mining, April 2022.

\bibitem{ackermann2012streamkm++}
M.~R. Ackermann, M.~M{\"a}rtens, C.~Raupach, K.~Swierkot, C.~Lammersen, and C.~Sohler, ``Streamkm++ a clustering algorithm for data streams,'' \emph{Journal of Experimental Algorithmics (JEA)}, vol.~17, pp. 2--1, 2012.

\bibitem{meesuksabai2011hue}
W.~Meesuksabai, T.~Kangkachit, and K.~Waiyamai, ``Hue-stream: Evolution-based clustering technique for heterogeneous data streams with uncertainty,'' in \emph{Advanced Data Mining and Applications: 7th International Conference, ADMA 2011, Beijing, China, December 17-19, 2011, Proceedings, Part II 7}.\hskip 1em plus 0.5em minus 0.4em\relax Springer, 2011, pp. 27--40.

\bibitem{breunig2000lof}
M.~M. Breunig, H.-P. Kriegel, R.~T. Ng, and J.~Sander, ``Lof: identifying density-based local outliers,'' in \emph{Proceedings of the 2000 ACM SIGMOD international conference on Management of data}, 2000, pp. 93--104.

\bibitem{pokrajac2007incremental}
D.~Pokrajac, A.~Lazarevic, and L.~J. Latecki, ``Incremental local outlier detection for data streams,'' in \emph{2007 IEEE symposium on computational intelligence and data mining}.\hskip 1em plus 0.5em minus 0.4em\relax IEEE, 2007, pp. 504--515.

\bibitem{salehi2016fast}
M.~Salehi, C.~Leckie, J.~C. Bezdek, T.~Vaithianathan, and X.~Zhang, ``Fast memory efficient local outlier detection in data streams,'' \emph{IEEE Transactions on Knowledge and Data Engineering}, vol.~28, no.~12, pp. 3246--3260, 2016.

\bibitem{ghorbani2019mahalanobis}
H.~Ghorbani, ``Mahalanobis distance and its application for detecting multivariate outliers,'' \emph{Facta Universitatis, Series: Mathematics and Informatics}, pp. 583--595, 2019.

\bibitem{gama2014survey}
J.~Gama, I.~{\v{Z}}liobait{\.e}, A.~Bifet, M.~Pechenizkiy, and A.~Bouchachia, ``A survey on concept drift adaptation,'' \emph{ACM computing surveys (CSUR)}, vol.~46, no.~4, pp. 1--37, 2014.

\end{thebibliography}

\end{document}